# A Novel Quadratic Interpolated Beetle Antennae Search for Manipulator Calibration

Zhibin Li, Shuai Li, *Senior Member*, *IEEE*, and Xin Luo, *Senior Member*, *IEEE*

*Abstract*—Over the past decades, industrial manipulators play a vital role in in various fields, like aircraft manufacturing and automobile manufacturing. However, an industrial manipulator without calibration suffers from its low absolute positioning accuracy, which extensively restricts its application in high-precision intelligent manufacture. Recent manipulator calibration methods are developed to address this issue, while they frequently encounter long-tail convergence and low calibration accuracy. To address this thorny issue, this work proposes a novel manipulator calibration method incorporating an extended Kalman filter with a Quadratic Interpolated Beetle Antennae Search algorithm. This paper has three-fold ideas: a) proposing a new Quadratic Interpolated Beetle Antennae Search algorithm to deal with the issue of local optimum and low convergence rate in a Beetle Antennae Search algorithm; b) adopting an extended Kalman filter algorithm to suppress non-Gaussian noises and c) developing a new manipulator calibration method incorporating an extended Kalman filter with a Quadratic Interpolated Beetle Antennae Search algorithm to calibrating a manipulator. Extensively experimental results on an ABB IRB120 industrial manipulator demonstrate that the proposed method achieves much higher calibration accuracy than several state-of-the-art calibration methods.

## I. INTRODUCTION

INDUSTRIAL manipulators achieve a rapid development in the Industry 4.0, which has been widely employed in intelligent manufacture, intelligent life, intelligent medicine and other advanced fields [1-5]. Generally, a manipulator enjoys its high repetitive positioning accuracy. However, it suffers from a low absolute positioning accuracy. Hence, it is of enormous significance to calibrate industrial manipulators [5-10].

Aiming at addressing this significant issue, numerous experts on manipulator calibration have conducted extensively related studies [12-15]. To date, various intelligent algorithms are proposed to search actual kinematic parameters, such as particle filter (PF) algorithm [5], Levenberg-Marquardt (LM) algorithm [9], support vector machine (SVM) algorithm [34] and neural networks [37]. Luo *et al*. [2] proposed a calibration method based on LM and differential evolution hybrid algorithm to estimate kinematic parameter deviations. Zhong *et al*. [4] utilized an improved whale swarm algorithm to calibrate the legged manipulators with a constraint handling mechanism, which successfully obtained kinematic parameter deviations. In general, the mentioned algorithms achieve a significantly improvement on the manipulator calibration accuracy [16], [17]. However, they frequently encounters slow convergence rate and low calibration accuracy [16-20].

In general, evolutionary computing algorithms are based on intuitive or empirical construction, which are adopted to address complex optimization issues. Traditional evolutionary computing algorithms include differential evolution algorithm (DE) [25], iterated greedy algorithm (IG) [42], particle swarm optimization algorithm (PSO) [36] and ant colony optimization algorithm (ACO). A Beetle Antennae Search (BAS) algorithm is one of evolutionary computing algorithms, which simulates foraging behavior of a beetle to search the optimal solution [41], [44]. Compared with other evolutionary computing algorithms, the search of a BAS algorithm only involves one individual, which has the advantages of ease of implementation, minimal computing resources and strong search ability [21-25].

For the above-mentioned virtues of a BAS algorithm, a novel calibration method with its principle is established. However, a BAS algorithm suffers from local optimum. To overcome the above-mentioned shortcoming of a BAS algorithm, we develop a new calibration method based on extended Kalman filter (EKF) algorithm and Quadratic Interpolated Beetle Antennae Search (QIBAS) algorithm. Moreover, considering the influence of non-Gaussian noises, the EKF algorithm effectively overcomes this issue [18], [20]. Enough experiments are conducted on ABB IRB120 manipulator to verify the feasibility of the proposed method [25-30]. The main contributions of this paper include:

a) Incorporating the quadratic interpolation into the updating rule of a BAS algorithm to obtain a QIBAS algorithm with fast with fast convergence and high calibration accuracy.
b) Employing an EKF algorithm to address the issue of non-Gaussian noises; and

✧ Z. Li and X. Luo are with the School of Computer Science and Technology, Chongqing University of Posts and Telecommunications, Chongqing 400065, China (e-mail: LiZhibin111@outlook.com, luoxin21@gmail.com).
✧ S. Li and X, Luo are with Zienkiewicz Centre for Computational Engineering and Department of Mechanical Engineering, College of Engineering, Swansea University Bay Campus, Swansea SA1 8EN, UK (email: shuai.li@swansea.ac.uk, luoxin21@gmail.com).

c) Developing an EKF-QIBAS approach to improve the calibration accuracy of a manipulator.

The rest structure of this paper is organized into four sections. Section II presents the kinematic model and error model. Section III describes the EKF and QIBAS algorithm for manipulator kinematic parameter identification. In Section IV, we discuss the experiments and comparison. Finally, Section V gives the conclusions.

## II. KINEMATIC MODEL AND ERROR MODEL FOR CALIBRATION

### A. Kinematic Model and Error Model

In general, the most commonly adopted kinematic model is the popular DH model proposed by Denavit and Hartenberg. According to the establishment principle of the DH model, the transformation relationship between the two adjacent link coordinate systems of the manipulator can be expressed as the following equation [36-42].

$$A_i = \begin{bmatrix} \cos\theta_i & -\sin\theta_i \cos\alpha_i & \sin\theta_i \sin\alpha_i & a_i \cos\theta_i \\ \sin\theta_i & \cos\theta_i \cos\alpha_i & -\cos\theta_i \sin\alpha_i & a_i \sin\theta_i \\ 0 & \sin\alpha_i & \cos\alpha_i & d_i \\ 0 & 0 & 0 & 1 \end{bmatrix}, \quad (1)$$

where $A_i$ represents the transformation matrix, $a_i$, $d_i$, $\theta_i$ and $\alpha_i$ are link length, link offset, joint angle, link twist angle of the $i$th link, respectively.

Then, the link transformation matrices are multiplied to achieve the manipulator kinematic model.

$$T_6 = A_1 A_2 A_3 A_4 A_5 A_6 = \begin{bmatrix} R_N & P_N \\ 0 & 1 \end{bmatrix}, \quad (2)$$

$R_N$, $P_N$ is the nominal rotation matrix and the nominal position vector, respectively.

According to equation (2), the pose error matrix of the manipulator end-effector is given as

$$dT = T_r - T_6, \quad (3)$$

where $dT$ is the pose deviation, $T_r$ is the actual pose of the manipulator end-effector, $T_6$ is the nominal pose of the manipulator end-effector.

We can obtain the holomorphic differential form of $A_i$.

$$dA_i = \frac{\partial A_i}{\partial \alpha_i} d\alpha_i + \frac{\partial A_i}{\partial a_i} da_i + \frac{\partial A_i}{\partial d_i} dd_i + \frac{\partial A_i}{\partial \theta_i} d\theta_i, \quad (4)$$

By combining (2) and (4), we achieve the pose error model of the manipulator.

$$\Delta E = \begin{bmatrix} H_1 & H_2 & H_3 & H_4 \end{bmatrix} \begin{bmatrix} \Delta a \\ \Delta d \\ \Delta \alpha \\ \Delta \theta \end{bmatrix} = H \Delta \eta, \quad (5)$$

$H$ is the extended Jacobian matrix, $\Delta \eta$ is the kinematic parameters deviations, which includes $\Delta\theta$, $\Delta d$, $\Delta a$, $\Delta\alpha$.

Based on the nominal cable length $Y_i'$ and the measuring cable length $Y_i$, the objective function for optimizing the kinematic parameter errors of the manipulator can be expressed as follow:

$$f = \min \|\Delta E\|_2^2 = \min\left[\frac{1}{n}\sum_{i=1}^{n}(Y_i - Y_i')^2\right], \quad (6)$$

In general, the nominal cable length can be calculated from the manipulator end-effector position coordinate $P_i$ and the fixed point coordinate $P_0$ on the ground, we have

$$Y_i' = \sqrt{(P_i - P_0)^2}. \quad (7)$$

## III. MANIPULATOR KINEMATIC PARAMETER IDENTIFICATION ALGORITHM

### A. QIBAS Algorithm

A BAS algorithm is also one of evolutionary computing algorithms, which is an intelligent optimization algorithm proposed by Jiang et al [17]. This method simulates the food searching behavior of a beetle.

Consider our objective function $f$ with the decision parameter being as follow:

$$\eta = [\eta_1, \eta_2, \cdots \eta_n]^T, \quad (8)$$

According to the principle of BAS algorithm, we give the following iterative model.

$$\eta_{t+1} = \eta_t + \delta_t \vec{b} \, sign(f(\eta_{rt}) - f(\eta_{lt})), \quad (9)$$

where $\delta$, $\vec{b}$ and $sign(\cdot)$ are the step size of searching, a random direction of beetle searching and the sign function, respectively. Then we obtain that

$$\vec{b} = \frac{rands(k,1)}{\|rands(k,1)\|_2}, \tag{10}$$

Evidently, rands and $k$ are the random function, the dimensions of position, respectively. $\|\cdot\|_2$ is the two-norm operator. Then, the searching behaviors of the left and right beetle's antennae are defined as

$$\eta_{rt} = \eta_t + m_t\vec{b},$$
$$\eta_{lt} = \eta_t - m_t\vec{b}, \tag{11}$$

where $m$ represents the sensing length of antennae. Then $m$ and step size of searching $\delta$ are updated as follow:

$$\delta_{t+1} = \mu\delta_t + \delta_0,$$
$$m_{t+1} = \tau m_t + m_0. \tag{12}$$

Generally, $\mu\in(0,1)$, $\tau\in(0,1)$, $\delta_0>0$, $m_0>0$. Note that, this two parameters $m$, $\delta$ can be set as constants to facilitate calculation if necessary. A BAS algorithm can be modified as

$$\eta_{t+1} = \eta_t + \delta_t\vec{b}\,sign(f(\eta_{rt}) - f(\eta_{lt})) \tag{13}$$

Commonly, due to BAS algorithm only has one individual, the ability of a beetle distinguishing the food odor concentration is not enough to find the optimal solution in the high-dimensional space. To address this issue, quadratic interpolation operator is incorporated into the updating rule of a BAS algorithm.

Assuming the approximate expression of the objective function is a second-order polynomial, we have

$$h(\eta_i) = c_0 + c_1\eta_i + c_2\eta_i^2 = f(\eta_i), \tag{14}$$

where $c_0$, $c_1$, $c_2$ are constants. To achieve the stable equilibrium point of binomial $h(\eta_i)$, whose first derivative is zero, we obtain that

$$h'(\eta_i) = c_1 + 2c_2\eta_i = 0, \Rightarrow \eta_{tk} = -\frac{c_1}{2c_2}, \tag{15}$$

Then, we achieve that

$$\begin{cases} h(\eta_{lk}) = c_0 + c_1\eta_{lk} + c_2\eta_{lk}^2 = f(\eta_{lk}) = f_1, \\ h(\eta_{rk}) = c_0 + c_1\eta_{rk} + c_2\eta_{rk}^2 = f(\eta_{rk}) = f_2, \Rightarrow \chi = \frac{(\chi_1^2 - \chi_3^2)f_2 + (\chi_3^2 - \chi_2^2)f_1 + (\chi_2^2 - \chi_1^2)f_3}{2((\chi_1 - \chi_3)f_2 + (\chi_3 - \chi_2)f_1 + (\chi_2 - \chi_1)f_3) + \upsilon_0}. \\ h(\eta_{bk}) = c_0 + c_1\eta_{bk} + c_2\eta_{bk}^2 = f(\eta_{bk}) = f_3. \end{cases} \tag{16}$$

where $k = 1, 2, \cdots, n$, $\chi_1 = \eta_{lk}$, $\chi_2 = \eta_{rk}$, $\chi_3 = \eta_{bk}$, $\chi = \eta_{tk}$, $f(\cdot)$ is the fitness function, $\upsilon_0$ is a small positive number.

### B. EKF Algorithm

Due to the effect of measurement noises, the EKF algorithm is applied to estimate the kinematic parameter errors of the manipulator, whose state equation and output equation are as follow:

$$\eta_{k|k-1} = \eta_{k-1|k-1}, Z_k = H_k\eta_k + V_k. \tag{17}$$

where $\eta$, $Z_k$, $H$ and $V_k$ are the deviations of the kinematic parameters, the end-effector's measurement position error, the Jacobian matrix and the measurement error, respectively.

Then we can achieve the error covariance matrix $P_k$ and the covariance matrix of the system noise $Q_k$.

$$P_{k|k-1} = P_{k-1|k-1} + Q_{k-1}, \tag{18}$$

By combining (17) and (18), the optimal Kalman gain $K_k$ is given as

$$K_k = P_{k|k-1}H_k^T\left(H_kP_{k|k-1}H_k^T + R_k\right)^{-1}, \tag{19}$$

where $R_k$ is the covariance matrix of $V_k$, then the recursive identification equation of $\eta$ can be expressed as the following equation

$$\eta_{k|k} = \eta_{k|k-1} + K_k\left(Z_k - H_k\eta_{k|k-1}\right), \tag{20}$$

The covariance matrix $P$ can be represented as

$$P_{k|k} = (I - K_kH_k)P_{k|k-1}. \tag{21}$$

## IV. EXPERIMENTAL RESULTS

### A. General Settings

*1) Evaluation Metrics:* In general, the root mean squared error (RMSE), average error (Std) and the maximum error (Max) have long been utilized as the evaluation metrics for the manipulator calibration method, which are summarized as follows:

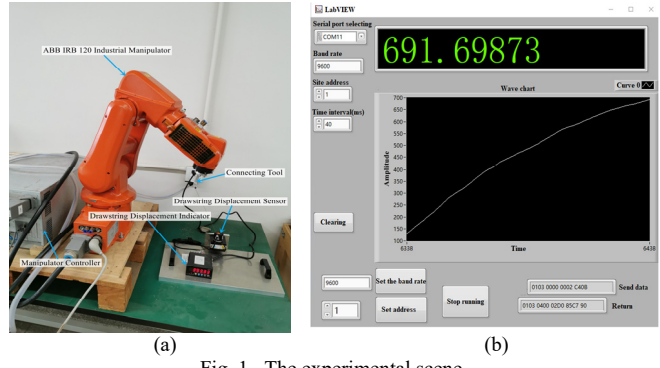

Fig. 1. The experimental scene.

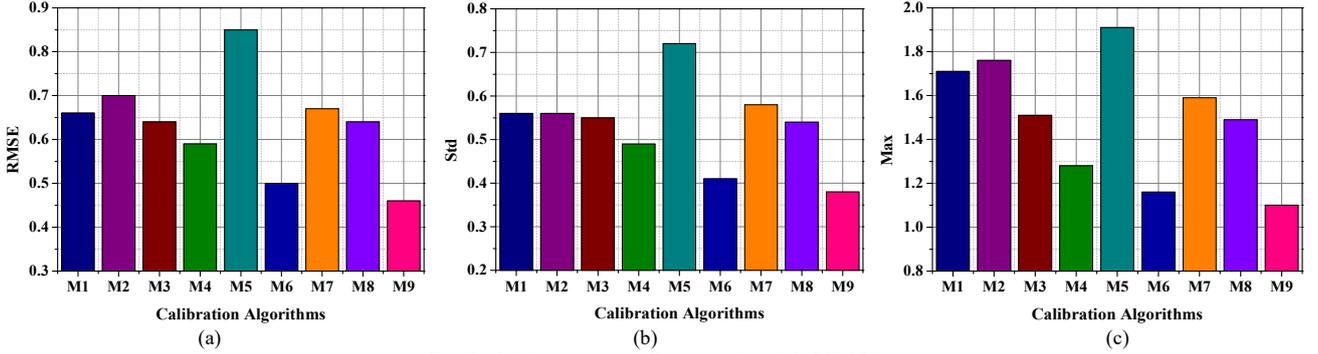

Fig. 2. Calibration error of compared models M1-M9.

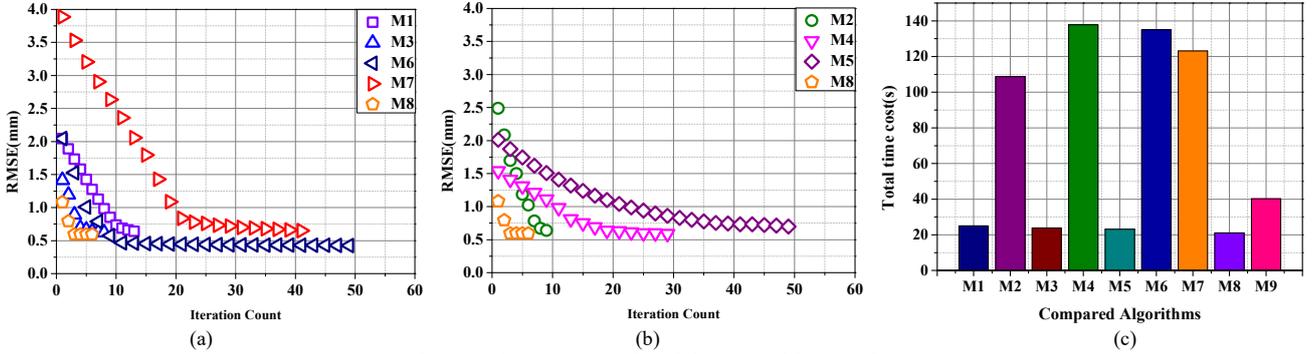

Fig. 3. Training curves and total time cost of these modes.

$$Max = \max\left\{\sqrt{(Y_i - Y_i')^2}\right\}, Std = \frac{1}{n}\sum_{i=1}^{n}\sqrt{(Y_i - Y_i')^2}, RMSE = \sqrt{\frac{1}{n}\sum_{i=1}^{n}(Y_i - Y_i')^2}, i = 1, 2, \cdots n \qquad (22)$$

*2) Dataset:* We adopt a drawstring displacement sensor to collect the position information of 120 measurement points on ABB IRB 120 industrial manipulator. Then, we adopt the 80%-20% train–test settings to achieve the objective experimental results [11].

*3) Experimental Setup:* We establish an experimental system, which is presented in Fig. 1, including a 6-DOF ABB IRB120 industrial manipulator, a drawstring displacement sensor, a drawstring displacement indicator, RS485 communication module.

*4) Experimental Process:* The positions of these measuring points are measured by a drawstring displacement sensor. The LabVIEW software is employed to develop a data collection software based on the RS485 communication module. Moreover, the developed data acquisition software is presented in Fig. 1(b). After finishing the measurement process, we obtain 120 samples.

*B. Comparison*

*1) Compared Methods*

In this part, we compare the proposed EKF-QIBAS method with several advanced manipulator calibration methods.

a) **M1: Extended Kalman filter (EKF) algorithm**, which is proposed in [16].
b) **M2: Beetle antennae search (BAS) algorithm** [19]. It simulates the food searching behavior of a beetle.
c) **M3: Unscented Kalman filter algorithm (UKF)**, which is presented in [42].

TABLE I
THE CALIBRATION ACCURACY OF COMPARED MODELS.

| Item | RMSE(mm) | Std(mm) | Max(mm) |
|---|---|---|---|
| Before | 2.09 | 2.00 | 3.36 |
| M1 | 0.66 | 0.56 | 1.71 |
| M2 | 0.70 | 0.56 | 1.76 |
| M3 | 0.64 | 0.55 | 1.51 |
| M4 | 0.59 | 0.49 | 1.28 |
| M5 | 0.85 | 0.72 | 1.91 |
| M6 | 0.50 | 0.41 | 1.16 |
| M7 | 0.67 | 0.58 | 1.59 |
| M8 | 0.64 | 0.54 | 1.49 |
| M9 | 0.46 | 0.38 | 1.10 |

d) **M4: Particle swarm optimization algorithm (PSO)** [23], which simulates the foraging behavior of a bird.

e) **M5: Radial basis function neural network (RBF)**, which is developed in [37].

f) **M6: Levenberg-Marquardt (LM) algorithm** [9], which is the most widely adopted nonlinear least squares algorithm.

g) **M7: Differential evolution algorithm (DE)** [25], which is proposed on the basis of the genetic algorithm.

h) **M8: Quadratic interpolated beetle antennae search (QIBAS) algorithm**, which greatly improves the global convergence ability of BAS algorithm.

i) **M9: EKF-QIBAS method**. A calibration method proposed in this work.

*2) Experimental Performance of these Methods*

Actually, the RMSE, Std and Max of the compared models are listed in Table I. Figs. 2 shows the calibration accuracy after calibration through these methods. Fig. 3 depicts their training curves and total time costs. From these experimental results, we can obtain the following findings.

a) M9 has the highest calibration accuracy among M1-M8. As depicted in Fig. 2, M9's RMSE, Std and Max are 0.46, 0.38 and 1.10, respectively. Compared with the most accurate single model M6, its RMSE, Std and Max are 0.50, 0.41 and 1.16 respectively, the accuracy gains are 8%, 7.32% and 5.17%, respectively. Hence, our proposed method is effective for improving manipulator calibration accuracy.

b) As shown in Fig. 3(a) and (b), M8 has a fastest convergence speed. M8 only takes 2 iterations to converge in RMSE. Compared with the fastest M2, which takes 7 iterations to converge in RMSE. From these experimental results, we can see that incorporating the quadratic interpolation into the updating rule of a BAS algorithm can improve its convergence speed.

c) As shown in Fig. 3(c), M8 has the highest computational efficiency than its peers do. It takes 21.04s for M8 to converge in RMSE, which is an improvement of 11.78% of 23.85s by the fastest M3. From the above results, we evidently see that it is greatly appropriate to incorporate quadratic interpolation into the updating rule of a BAS algorithm.

V. CONCLUSIONS

To accurately identify the errors of manipulator kinematic parameters, EKF-QIBAS method is proposed in this work. To our best knowledge, compared with the existing methods, it has a higher calibration accuracy than its peers do.